# SAMSNeRF: Segment Anything Model (SAM) Guides Dynamic Surgical Scene Reconstruction by Neural Radiance Field (NeRF)

Ange Lou[1], Yamin Li[2], Xing Yao[2], Yike Zhang[2], Jack Noble[1]

Department of Electrical Engineering[1], Vanderbilt University, Nashville TN, USA
Department of Computer Science[2], Vanderbilt University, Nashville TN, USA
{ange.lou, yamin.li, xing.yao, yike.zhang, jack.noble}@vanderbilt.edu

## Abstract

The accurate reconstruction of surgical scenes from surgical videos is critical for various applications, including intraoperative navigation and image-guided robotic surgery automation. However, previous approaches, mainly relying on depth estimation, have limited effectiveness in reconstructing surgical scenes with moving surgical tools. To address this limitation and provide accurate 3D position prediction for surgical tools in all frames, we propose a novel approach called **SAMSNeRF** that combines Segment Anything Model (SAM) and Neural Radiance Field (NeRF) techniques. Our approach generates accurate segmentation masks of surgical tools using SAM, which guides the refinement of the dynamic surgical scene reconstruction by NeRF. Our experimental results on public endoscopy surgical videos demonstrate that our approach successfully reconstructs high-fidelity dynamic surgical scenes and accurately reflects the spatial information of surgical tools. Our proposed approach can significantly enhance surgical navigation and automation by providing surgeons with accurate 3D position information of surgical tools during surgery. The code will be released soon at: https://github.com/AngeLouCN/SAMSNeRF

**Keywords:** Segment Anything Model, Neural Radiance Field, Surgical scene reconstruction

## 1. INTRODUCTION

Deep learning has exerted a profound influence on the comprehension of surgical scenes, particularly in tasks such as tracking surgical tools within 2D videos through segmentation and detection algorithms, surgical phase recognition, reconstruction [13], and depth estimation [14]. In these domains, 2D-level algorithms can only provide the surgeon with 2D positional information, presenting limitations in the context of real-time image-guided surgery. Consider cochlear implant surgery, for instance, where obtaining 3D position and pose information for surgical insertion tools is critical for precise treatment by aligning with pre-operatively planned insertion trajectories [15]. The integration of 3D reconstruction techniques into the surgical field has gained significant attention and holds promise for enhancing surgical procedures.

Reconstructing surgical scenes from surgical videos is a crucial but challenging task in minimally invasive surgery [1]. It serves as a prerequisite for many downstream clinical applications, including navigation, augmented reality, and surgical education [2]. Previous works [3][4] have shown the effectiveness of surgical scene reconstruction via depth estimation. However, these methods work well only in static scenes, and they become unreliable when there are moving objects such as surgical tools in the scene. Moreover, these methods reconstruct the scene on a sparse warp field, which causes the loss of texture information between pixels. Recently, neural rendering has overcome the limitations of traditional 3D reconstruction by using differential rendering and neural networks. The Neural Radiance Field (NeRF) [5] is a popular approach that synthesizes high-fidelity novel views of static scenes by using neural implicit fields to represent continuous scene representations. Moreover, variants of NeRF [6][7] have been developed to solve the dynamic scene reconstruction problem and achieve promising results. However, the presence of moving objects in surgical scenes leads to uncertainties that must be optimized. To address these uncertainties, segmentation networks [8][9] can be used to filter the background and foreground in pixel-level detail. We apply the Segment Anything Model (SAM) [10] to segment the surgical tool using a rough bounding box as input prompt, which provides a more accurate segmentation mask compared to some other CNN-based methods [11]. Our goal in this paper is to build a model to reconstruct the surgical scene and track the movement of surgical tools. Our contributions are:

(1) To the best of our knowledge, SAMSNeRF is the first method that can reconstruct the surgical scene with surgical tools.
(2) For the first time, we combine the powerful vision model, SAM, with NeRF to achieve more accurate reconstruction of surgical scenes.

## 2. METHOD

### 2.1. Overview of SAMSNeRF framework.

In our SAMSNeRF framework, our goal is to reconstruct 3D structures and textures of surgical scenes from a given surgical video. To achieve this, we first obtain the input video frames denoted as $\{I_i\}_{i=1}^{T}$, where $T$ is the number of frames and the duration of the video is normalized to $[0,1]$ with time of the $i$-th frame being $i/T$. We also use SAM to obtain surgical tool masks $\{M_i\}_{i=1}^{T}$ to identify the regions of surgical tools in the video frames. Next, we use an existing pretrained depth estimation network to obtain coarse depth maps for the video, which are denoted as $\{D_i\}_{i=1}^{T}$. With these prerequisites, we follow the modeling in EndoNeRF [7] and represent the dynamic surgical scenes as a canonical neural radiance field along with a time-dependent neural displacement field. The overview architecture of our SAMSNeRF approach is illustrated in Figure 1.

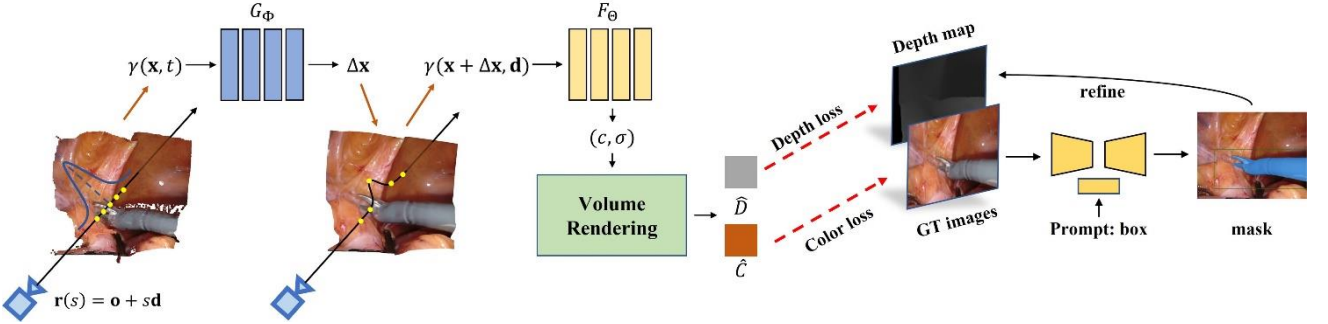

Figure 1. Architecture of SAMSNeRF. We proposed SAM based refinement method to reconstruct surgical scene.

### 2.2. EndoNeRF.

The surgical scenes are represented by a canonical radiance field and a time-dependent displacement field. The canonical field is represented by an 8-layer MLP neural network, denoted as $F_\Theta(\mathbf{x}, \mathbf{d})$. It maps the 3D coordinates $\mathbf{x} \in \mathbb{R}^3$ and viewpoint direction $\mathbf{d} \in \mathbb{R}^3$ to RGB colors $c(\mathbf{x}, \mathbf{d}) \in \mathbb{R}^3$ and space occupancy $\sigma(\mathbf{x}) \in \mathbb{R}$. The time-dependent displacement field is represented by another 8-layer MLP, denoted as $G_\Phi(\mathbf{x}, t)$. It maps the input space-time coordinates $(\mathbf{x}, t)$ to the displacement between the point $x$ at time $t$ and corresponding point in canonical field. At any given time $t$, the color and occupancy at point $x$ can be retrieved as $F_\Theta(\mathbf{x} + G_\Phi(\mathbf{x}, t), \mathbf{d})$. To capture high-frequency information, we also use a position encoding $\gamma(\cdot)$ to map the points' coordinates and time. After obtaining the scene representation, we leverage the differential volume rendering to generate renderings (depth and rendered images) for supervision. The differential volume rendering begins with shooting a batch of camera rays into the surgical scene from a fixed viewpoint at an arbitrary time $t$. Each ray is formulated as $\mathbf{r}(s) = \mathbf{o} + s\mathbf{d}$, where $\mathbf{o}$ is a fixed origin of a single ray, $\mathbf{d}$ is the direction of the ray and $s$ is the ray parameter. Once the sample points are obtained, the emitted color $\hat{C}$ and depth $\hat{D}$ of a ray $\mathbf{r}(s)$ can be evaluated by volume rendering as:

$$\hat{C}(\mathbf{r}(s)) = \sum_{j=1}^{m-1} w_j c(\mathbf{x}_j, \mathbf{d}), \qquad \hat{D}(\mathbf{r}(s)) = \sum_{j=1}^{m-1} w_j s_j,$$

$$w_j = \left(1 - \exp(-\sigma(\mathbf{x}_j)\Delta s_j)\right) \exp\left(-\sum_{k=1}^{j-1} \sigma(\mathbf{x}_k)\Delta s_k\right), \qquad \Delta s_j = s_{j+1} - s_j. \tag{1}$$

The networks are optimized by rendered color and depth, the loss function is defined as:

$$\mathcal{L}(\mathbf{r}(s)) = \left\|\hat{C}(\mathbf{r}(s)) - I[u,v]\right\|_2^2 + \left|\hat{D}(\mathbf{r}(s)) - D[u,v]\right|, \tag{2}$$

where $(u, v)$ is the location of the pixel.

### 2.3. Segment Anything Model.

The Segment Anything Model (SAM) is a powerful foundation model for image segmentation that has been extensively trained on a vast dataset of over 1 billion masks from 11 million licensed and privacy-respecting images. This extensive training has enabled SAM to achieve state-of-the-art performance in image segmentation tasks. One of the key advantages of SAM is its ability to support zero-shot image segmentation with various prompts, including points, boxes, masks, and text. This makes it possible to perform image segmentation on medical or surgical images without the need for additional fine tuning. To guide the SAM in segmenting the tools accurately, we utilize manually selected boxes as prompts. These boxes are chosen based on the knowledge that the tools move within specific areas during the surgery. And the $i$-th frame's tool mask is denoted as $M_i$.

**2.4. Depth refinement.**

In our approach, we proposed depth refinement for both the background tissues and foreground surgical tools due to the presence of specular reflection and fuzzy pixels in the depth maps. At the early training stage, the network tends to produce smoother results in depth due to underfitting, where the model tends to average learned colors and occupancy. This results in the majority of normal depths smoothing out some minority corrupt depths. Denote the predicted depth for $i$-th frame after $K$ iteration training as $\widehat{D}_i^K$, we find the residual maps of background tissues as $\epsilon_i^B = |\widehat{D}_i^K - D_i| \odot (1 - M_i)$, where $M_i$ is the binary mask of the foreground surgical tools. We then calculate the probabilistic distribution over the residual maps and set a small number $\alpha$, which is in the range [0, 1]. We locate those pixels with the last $\alpha$-quantile residuals, which correspond to the pixels with large residuals, and replace those depth pixels with smooth background depth pixels in $\widehat{D}_i^K$. For the foreground surgical tools, we define the residual maps as $\epsilon_i^F = |\widehat{D}_i^K - D_i| \odot M_i$. We also replace the last $\alpha$-quantile pixels with smooth foreground depth pixels to refine the depth for the surgical tools.

## 3. EXPERIMENTS

**3.1. Dataset.**

We utilize the EndoNeRF dataset, comprising two video cases consisting of 63 and 156 frames each. These cases are recorded using stereo cameras from a fixed viewpoint. To assess the quality of our reconstruction, we conduct a qualitative evaluation by visualizing the reconstructed point cloud and comparing it with the EndoNeRF. Furthermore, we conducted a comparative analysis to assess the impact of segmentation masks obtained from different methods – CNN [8], SAM, and ground truth. We use the PSNR, SSIM and LPIPS, as evaluation metrics for quantitative comparisons.

**3.2. Implementation details.**

In all experiments, we have fixed the depth refinement iteration at $K = 4000$, and the remaining training hyperparameters have been set to match those in D-NeRF [6]. For generation of coarse depth maps, we utilized STTR-light [12] pretrained on Scene Flow. Each case is trained for $100K$ iterations on a single RTX A5000 GPU about 12 hours.

**3.3. Results**

In Table 1, we present the mean PSNR, SSIM, and LPIPS [16] scores for two cases in the EndoNeRF dataset. Prior to introducing our foreground and background refinement techniques, the reconstruction of the high-fidelity surgical scene, inclusive of surgical instruments, was unsuccessful. However, upon implementing both foreground and background refinements, a substantial improvement in PSNR from 21.435 to 32.127 was observed. Notably, when utilizing a more precise segmentation mask as shown in Figure 2 to guide the refinement process, the accuracy further improved.

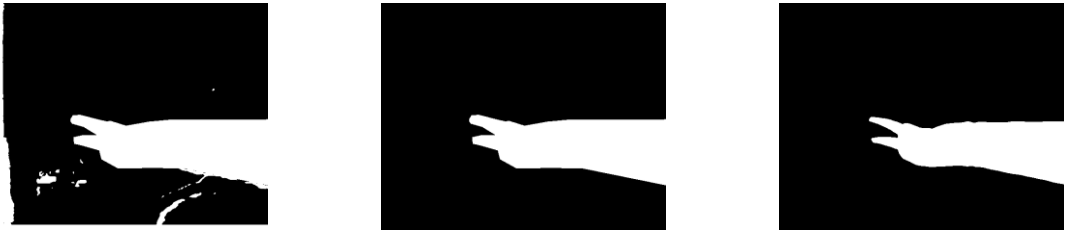

Figure 2. From left to right are segmentation masks generated by CNN, manually and SAM.

Table 1. Quantitative result of scene reconstruction. "↑" represents higher is better and "↓" represents lower is better.

|  | PSNR ↑ | SSIM ↑ | LPIPS ↓ |
|---|---|---|---|
| EndoNeRF | 21.435 | 0.720 | 0.287 |
| ours w/ CNN | 32.127 | 0.904 | 0.112 |
| ours w/GT | 34.425 | 0.918 | 0.100 |
| ours w/ SAM (SAMSNeRF) | **34.537** | **0.921** | **0.095** |

Figure 3 (a) and (b) display the reconstruction quality achieved by the depth map and our SAMSNeRF, respectively. To facilitate a clear comparison, we visualize the point cloud from a left-side viewpoint, demonstrating that our SAMSNeRF outperforms the depth map by retaining rich texture information in a denser point cloud representation. In (c) and (d), we present the results obtained from EndoNeRF and SAMSNeRF, respectively. Leveraging both background and foreground refinement strategies, SAMSNeRF demonstrates its capability to successfully reconstruct the high-fidelity surgical scene, including the intricate details of surgical tools.

## 4. DISCUSSION

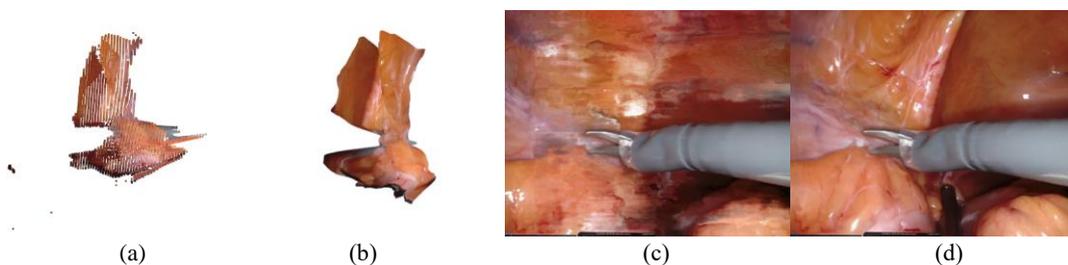

(a)          (b)          (c)          (d)

Figure 3. (a) and (b) are the 3D reconstruction results from depth map and SAMSNeRF. (c) and (d) are the reconstruction results of EndoNeRF and SAMSNeRF.

Our proposed background and foreground refinement strategy has been remarkably successful in reconstructing dynamic surgical scenes with surgical instruments. Furthermore, we have innovatively incorporated a "segment anything" model, enabling accurate segmentation mask predictions, which, in turn, guide the precise refinement of depth maps.

## 5. CONCLUSION

Our proposed SAMSNeRF have demonstrated impressive success in reconstructing surgical scenes with precision, incorporating the presence of surgical tools. However, training a NeRF model for a single case demands 12 hours. Additionally, the current requirement for a pre-existing depth map of the surgical scene prior to training is an additional limitation. Our future research endeavors will primarily focus on developing an efficient dynamic NeRF approach [17], aiming to significantly reduce training time while maintaining or even enhancing reconstruction accuracy. Simultaneously, we will explore and develop zero-shot depth estimation techniques to enable our proposed method to handle monocular dynamic scene reconstruction effectively [18]. Furthermore, the application of recent adaptor techniques [19] or post-processing [20] strategies can facilitate the transfer of SAM from regular image processing to the domain of surgical imagery. This adaptation is expected to enhance the robustness of segmentation performance in surgical videos to further refine the depth maps in our task.